\documentclass[11pt]{article}
\usepackage{eacl2017}
\usepackage{times}
\usepackage{url}
\usepackage{latexsym}
\usepackage[dvipdf]{graphicx}
\usepackage[dvipdfmx]{hyperref}

\eaclfinalcopy 
\def\eaclpaperid{44} 

\title{Bib2vec: An Embedding-based Search System for Bibliographic Information}

\author{Takuma Yoneda \qquad Koki Mori \qquad Makoto Miwa \qquad Yutaka Sasaki \\
Computational Intelligence Laboratory \\
Toyota Technological Institute \\
2-12-1 Hisakata, Tempaku-ku, Nagoya, Japan \\
  {\tt \{sd14084,sd15435,makoto-miwa,yutaka.sasaki\}@toyota-ti.ac.jp} \\}

\date{}

\begin{document}
\maketitle
\begin{abstract}
We propose a novel embedding model that represents relationships among several elements in bibliographic information with high representation ability and flexibility. Based on this model, we present a novel search system that shows the relationships among the elements in the ACL Anthology Reference Corpus. The evaluation results show that our model can achieve a high prediction ability and produce reasonable search results.
The demonstration is available at \href{http://tti-coin.jp/demo/bib2vec/}{http://tti-coin.jp/demo/bib2vec/}
\end{abstract}

\section{Introduction}

Modeling relationships among several types of information, such as nodes in information networks, has attracted great interests in natural language processing (NLP) and data mining (DM), since it can uncover hidden information in data. Topic models such as author-topic model~\cite{Rozen-Zvi} have been widely studied to represent relationships among these types of information. These models, however, need a considerable effort to incorporate new types and do not scale well in increasing the number of types since they explicitly model the relationships between types in the generating process.

Word representation models, such as skip-gram and continuous bag-of-word (CBOW) models ~\cite{T.Mikolov}, have made a great success in NLP. They have been widely used to represent texts, but recent studies started to apply these methods to represent other types of information, e.g., authors or papers in citation networks~\cite{line-large-scale-information-network-embedding}. 

We propose a novel embedding model that represents relationships among several elements in bibliographic information, which is useful to discover hidden relationships such as authors' interests and similar authors. We built a novel search system that enables to search for authors and words related to other authors based on the model using the ACL Anthology Reference Corpus~\cite{bird2008acl}. 
Based on skip-gram and CBOW models, our system embeds vectors to not only words but also other elements of bibliographic information such as authors and references and provides a great representation ability and flexibility.
The vectors can be used to calculate distances among the elements using similarity measures such as the cosine distance and inner products. For example, the distances can be used to find words or authors related to a specific author. 
Our model can easily incorporate new types without changing the model structure and scale well in the number of types. 

\section{Related work}

Most of previous studies on modeling several elements in bibliographic information have been based on topic models such as author-topic model~\cite{Rozen-Zvi}. Although the models work fairly well, they have comparably low flexibility and scalability since they explicitly model the generation process. Our model employs word representation-based models instead of topic models.

Some previous studies embedded vectors to the elements. Among them, large-scale information network embedding (LINE)~\cite{line-large-scale-information-network-embedding} embedded a vector to each node in information network. 
LINE handles single type of information and prepares a network for each element separately. By contrast, our model simultaneously handles all the types of information.
 
\section{Method}
We propose a novel method to represent bibliographic information by embedding vectors to elements based on the
skip-gram and CBOW models.

\subsection{Task definition}

We assume the bibliographic data set have the following structure.
The data set is composed of bibliographic information of papers.
Each paper consists of several categories. 
Categories are divided into two groups: a textual category \(\Psi\) (e.g., titles and abstracts{\footnote{Note that we have only one textual category since the categories for texts are usually not distinguished in most word representation models.}}) and non-textual categories \(\Phi\) (e.g., authors and references). Figure~\ref{fig:one} illustrates an example structure of bibliographic information of a paper. 
Each category has one or more elements; the textual category usually has many elements while a non-textual category has a few elements (e.g., authors are not many for a paper). 

\subsection{Proposed model} 

\begin{figure}[t!]
\centering
\includegraphics[width=.85\linewidth, clip=true, trim=0 2 0 0]{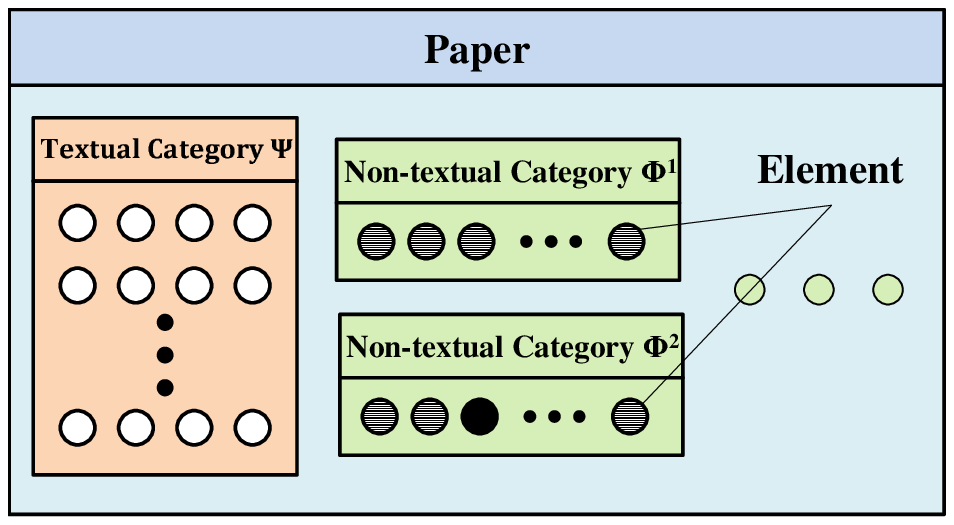}
\caption{Example of the bibliographic information of a paper 
when the target is the element in the non-textual category.
The black element is a target and the shaded elements are contexts.}
\label{fig:one}
\centering
\label{fig:two}
\includegraphics[width=.85\linewidth, clip=true, trim=0 5 0 0]{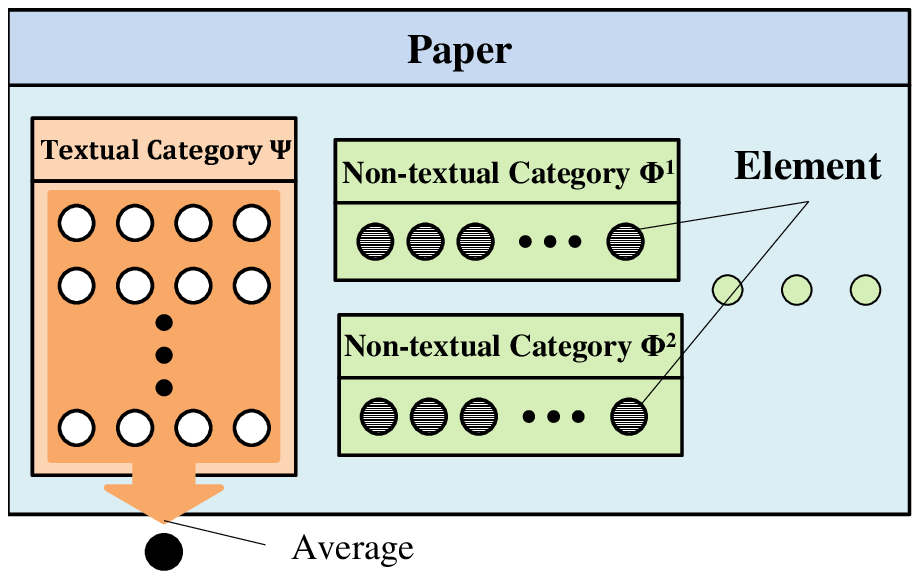}
\caption{Example when the target is the elements in the textual category}
\end{figure}

Our model focuses on a \textit{target} element, and predicts a \textit{context} element from the target element. We use only the elements in non-textual categories as contexts to reduce the computational cost.
Figure~\ref{fig:one} shows the case when we use an element in a non-textual category as a target. For the black-painted target element in category \({\Phi}^2\), the shaded elements in the same paper are used as its contexts. 

When we use elements in the textual category as a target, instead of treating each element as a target, we consider that the textual category has only one element that represents all the elements in the category like CBOW. Figure~\ref{fig:two} illustrates the case that we consider the averaged vector of the vectors of all the elements in the textual category as a target. 

We describe our probabilistic model to predict a context element \(e^{j}_{O}\) from a target \(e^{i}_{I}\) in a certain paper.
We define two $d$-dimensional vectors \(\upsilon^{i}_{t}\) and \(\omega^{i}_{t}\) to represent an element \(e^{i}_{t}\) as a target and context, respectively. 
Similarly to the skip-gram model, the probability to predict element \(e^{j}_{O}\) in the context from input \(e^{i}_{I}\) is defined as follows:
\begin{eqnarray}
p(e^{j}_{O}|e^{i}_{I})&=&\frac{\exp(\omega^{j}_{O}{\cdot}{\upsilon^{i}_{I}} + \beta^{j}_{O})}{\sum_{(\omega^{j}_{s},\beta^{j}_{s}) \in S^{j}}{\exp(\omega^{j}_{s}{\cdot}{\upsilon^{i}_{I}} + \beta^{j}_{s})}},\nonumber\\
&&e^{j}_{O} \in \Phi, \ \ e^{i}_{I} \in \Psi \cup \Phi,\label{eq:one}
\end{eqnarray}
where \( \beta^{j}_{s} \) denotes a bias corresponds to \( \omega^{j}_{s} \), and \(S^{j}\) denotes pairs of \(\omega^{j}_{s}\) and \(\beta^{j}_{s}\) that belong to a category \(\Phi^{j}\).
As we mentioned, our model considers that the textual category \(\Psi\) has only one averaged vector. The vector \(\upsilon_{rep}^{j}\) can be described as:
\begin{equation}
\upsilon_{rep}^{j} = \frac{1}{n} \sum_{q = 1}^{n} \upsilon_{q}^{j}, \  e^{j} \in \Psi
\end{equation}

Our target loss can be defined as:
\begin{equation}
-\sum_{(e_{a},e_{b})\in D}\log{p(e_{b}|e_{a})}, 
\end{equation}
where \(D\) denotes a set of all the correct pairs of the elements in the data set.
To reduce the cost of the summation in Eq.~(\ref{eq:one}), we applied the noise-contrastive estimation (NCE) to minimize the loss \cite{Gutmann2010}.

\subsection{Predicting related elements}

We predict the top $k$ elements related to a query element by calculating their similarities to the query element.  
We calculate the similarities using one of three similarity measures: the linear function in Eq.~(\ref{eq:one}), dot product, and cosine distance.

\section{Experiments}

\subsection{Evaluation settings}

\begin{table}[t]
\centering
\footnotesize
  \begin{tabular}{llrrr} \hline
  	& & \multicolumn{2}{c}{\#Elements} & Min.\\
    Category & Type & Original & Processed & Freq. \\ \hline
    text & textual & 59,276 & 10,994  & 20\\
    author & non-textual & 17,260 & 2,609 & 5\\
    reference & non-textual & 10,871 & 10,871 & 1\\
    year & non-textual &16 & 16 & 1\\
    paper-id & non-textual & 19,475 & 19,475 & 1\\ \hline
  \end{tabular}
  \caption{Summary of our data set and model}
 \label{tb:one}
\end{table}

We built our data set from the ACL Anthology Reference Corpus version 20160301~\cite{bird2008acl}.
The statistics of the data set and our model settings are summarized in Table~\ref{tb:one}.

As pre-processing, we deleted commas and periods that sticked to the tails of words and removed non-alphabetical words such as numbers and brackets from abstracts and titles.
We then lowercased the words, and made phrases using the word2phrase tool\footnote{\url{https://github.com/tmikolov/word2vec}}. 

We prepared five categories: \textit{author}, \textit{paper-id}, \textit{reference}, \textit{year} and \textit{text}.
\textit{author} consists of the list of authors without distinguishing the order of the authors.
\textit{paper-id} is an unique identifier assigned to each paper, and this mimics the paragraph vector model~\cite{pv}. 
\textit{reference} includes the paper ids of reference papers in this data set. Although ids in paper-id and reference are shared, we did not assign the same vectors to the ids since they are different categories.
\textit{year} is the publication year of the paper.
\textit{text} includes words and phrases in both abstracts and titles, and it belongs to the textual category \(\Psi\), while each other category is treated as a non-textual category \(\Phi^i\).
We regard elements as unknown elements when they appear less than minimum frequencies in Table~\ref{tb:one}.

We split the data set into training and test. We prepared 17,475 papers for training and the remaining 2,000 papers for evaluation. For the test set, we regarded the elements that do not appear in the training set as unknown elements.

We set the dimension $d$ of vectors to 300 and show the results with the linear function.

\begin{table*}[t!]
\footnotesize
\centering
  \begin{tabular}{lllll} \hline
 & \multicolumn{2}{c}{Our Model} & \multicolumn{2}{c}{Author Topic-Model} \\
Input Author & Relevant Words & Similar Authors & Topic Words & Topic Authors \\ \hline \hline
Philipp Koehn & machine translation & Hieu Hoang & alignment & Chris Dyer \\
 & hmeant & Alexandra Birch & translation & Qun Liu \\
 & human translators &Eva Hasler & align & Hermann Ney \\ \hline
 
Ryan McDonald & dependency parsing &Keith Hall & parse & Michael Collins \\
 & extrinsic & Slav Petrov & sentense & Joakim Nivre \\
 & hearing  & David Talbot& parser & Jens Nilson \\ \hline

  \end{tabular}
  \caption{Working examples of our model and author topic-model}
\label{tb:three}
\end{table*}
\subsection{Evaluation}
\begin{figure*}[t!]
\centering

\includegraphics[width=\linewidth,clip=false]{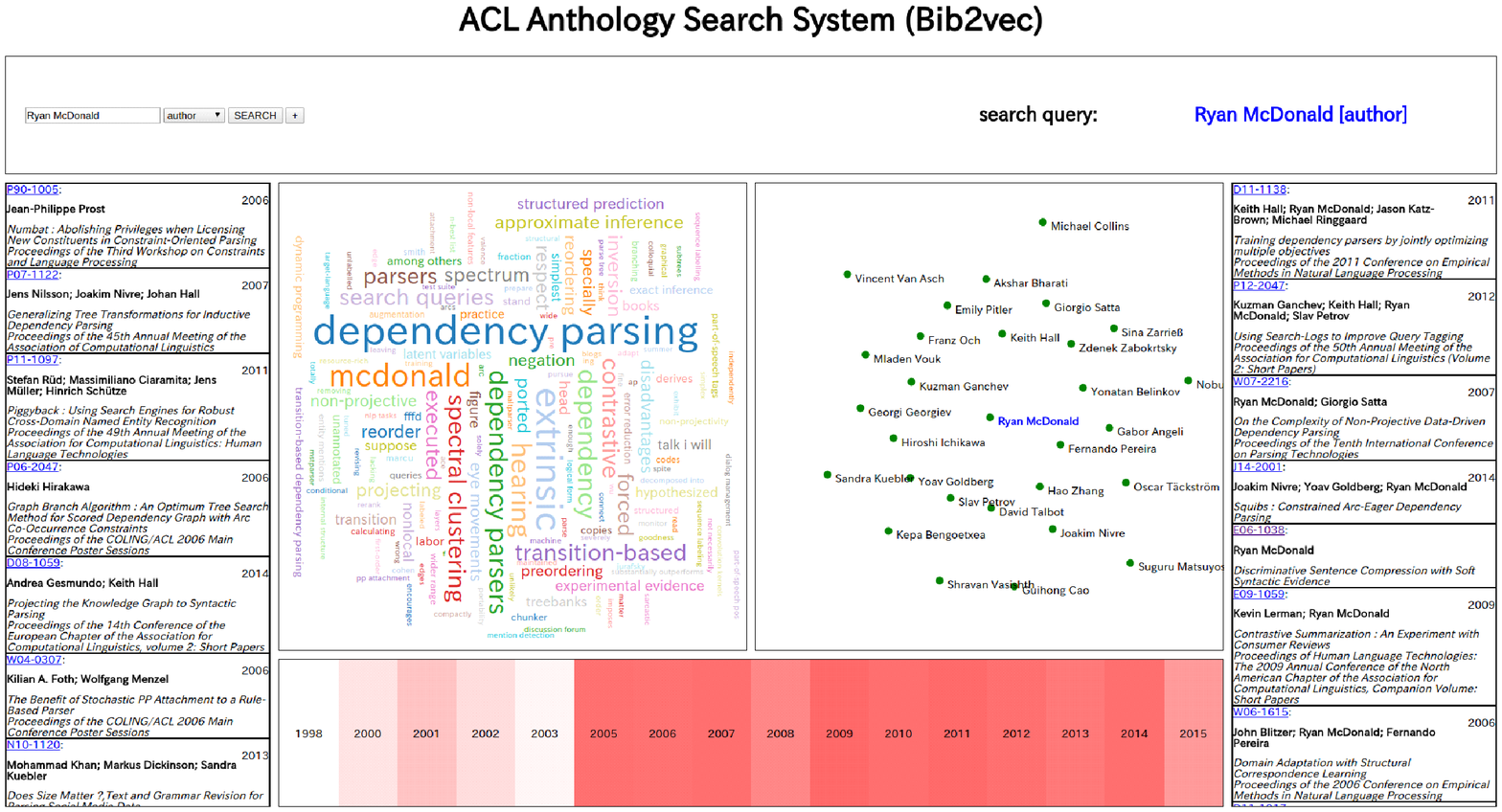}
\caption{Screen shot of the system with the search results for the query ``Ryan McDonald''.}
\label{fig:three}
\end{figure*}

We automatically built multiple choice questions and evaluate the accuracy of our model. 
We also compared some results of our model with those of author-topic model.

Our method models elements in several categories and allows us to estimate relationships among the elements with high flexibility, but this makes the evaluation complex. Since it is tough to evaluate all the possible combinations of inputs and targets, we focused on relationships between authors and other categories. 
We prepared an evaluation data set that requires to estimate an author from other elements. We removed an (not unknown) author from each paper in the evaluation set to ask the system to predict the removed author considering all the other elements in the paper.
To choose a correct author from all the authors can be insanely difficult, so we prepared 10 selection candidates.
In order to evaluate the effectiveness of our model, we compared the accuracy on this data set with that by logistic regression.
As a result, when we use our model, we got 74.3\% (1,486 / 2,000) in accuracy, which was comparable to 74.1\% (1,482 / 2,000) by logistic regression.

Table~\ref{tb:three} shows the examples of the search results using our model.
The leftmost column shows the authors we input to our model.
In the rightmost two columns, we manually picked up words and authors belonging to a certain topic described in \newcite{Yanchuan} that can be considered to correspond to the input author.
This table shows that our model can predict relative words or similar authors favorably  well although the words are inconsistent with those by the author topic model. 

Figure~\ref{fig:three} shows the screenshot of our system. The lefthand box shows words in the word cloud related to the query and the righthand box shows the close authors.
We can input a query by putting it in the textbox or click one of the authors in the righthand box and select a similarity measure by selecting a radio button. 

\subsection{Discussion}

When we train the model, we did not use elements in category \(\Psi\) as context. This reduced the computational costs, but this might disturbed the accuracy of the embeddings.
Furthermore, we used the averaged vector for the textual category \(\Psi\), so we do not consider the importance of each word. 
Our model might ignore the inter-dependency among elements since we applied skip-grams. 
To resolve these problems, we plan to incorporate attentions~\cite{cbowa} so that the model can pay more attentions to certain elements that are important to predict other elements.

We also found that some elements have several aspects. For example, words related to an author spread over several different tasks in NLP. We may be able to model this by embedding multiple vectors~\cite{Neelakantan}.

\section{Conclusions}

This paper proposed a novel embedding method that represents several elements in bibliographic information with high representation ability and flexibility, and presented a system that can search for relationships among the elements in the bibliographic information.
Experimental results in Table~\ref{tb:three} show that our model can predict relative words or similar authors favorably well.
We plan to extend our model by other modifications such as incorporating attention and embedding multiple vectors to an element. Since this model has high flexibility and scalability, it can be applied to not only papers but also a variety of bibliographic information in broad fields.


\section*{Acknowledgments}
We would like to thank the anonymous reviewer for helpful comments
and suggestions.

\bibliography{eacl2017}
\bibliographystyle{eacl2017}

\end{document}